\documentclass{article}

\usepackage[final,nonatbib]{nips_2018}

\usepackage[utf8]{inputenc} % allow utf-8 input
\usepackage[T1]{fontenc}    % use 8-bit T1 fonts
\usepackage{hyperref}       % hyperlinks
\usepackage{url}            % simple URL typesetting
\usepackage{booktabs}       % professional-quality tables
\usepackage{amsfonts}       % blackboard math symbols
\usepackage{nicefrac}       % compact symbols for 1/2, etc.
\usepackage{microtype}      % microtypography
\usepackage{amsmath}
\usepackage{amssymb}
\usepackage{mathtools}
\usepackage{bbm}
\usepackage{color}
\usepackage{cancel}
\usepackage{listings}
\usepackage{caption}
\usepackage{float}
\usepackage{todonotes}
\usepackage{ulem} % added by Hirofumi K.
\usepackage{graphicx}
\newcommand{\mathbbE}{\mathbb{E}}

\newcommand{\mathbbR}{\mathbb{R}}

\newcommand{\boldone}{{\boldsymbol{1}}}

\newcommand{\boldH}{{\boldsymbol{H}}}
\newcommand{\boldI}{{\boldsymbol{I}}}

\newcommand{\boldK}{{\boldsymbol{K}}}
\newcommand{\boldL}{{\boldsymbol{L}}}

\newcommand{\boldx}{{\boldsymbol{x}}}
\newcommand{\boldy}{{\boldsymbol{y}}}

\newcommand{\calL}{{\mathcal{L}}}

\newcommand{\calS}{{\mathcal{S}}}

\title{Modeling the Biological Pathology Continuum with HSIC-regularized Wasserstein Auto-encoders}

\author{
  Denny Wu \\
  University of Toronto, Vector Institute\\
\small{\texttt{dennywu@cs.toronto.edu}} \\
  \And
   Hirofumi Kobayashi \\
   University of Tokyo \\
 \small{\texttt{liamiiliil@chem.s.u-tokyo.ac.jp}} \\
   \And
   Charles Ding \\
   University of Toronto \\
\small{\texttt{charles.ding@mail.utoronto.ca}} \\
   \And
   Lei Cheng \\
   University of Tokyo \\
   \small{\texttt{leicheng@chem.s.utokyo.ac.jp}} \\
   \And
   Keisuke Goda \\
   University of Tokyo \\
   \small{\texttt{goda@chem.s.u-tokyo.ac.jp}} \\
   \And
  Marzyeh Ghassemi \\
  University of Toronto, Vector Institute\\
  \small{\texttt{marzyeh@cs.toronto.edu}} \\
}

\begin{document}


\maketitle

\begin{abstract}

% With proven effectiveness of deep learning models in classification tasks on biological data with discrete different morphology, the continuous changes in morphology remains unelucidated. We proposed to use generative model, in particular, HSIC-regularized Wasserstein Auto-encoders, to capture the overall representation of the pathology continuum within the context of biology. We further showed that, by using Hilbert-Schmidt independence criterion (HSIC) as a regularizer, we could also correlate the morphological changes with monotonic increasing factors. In this study, we successfully showed and validated its effectiveness on two distinct dataset with different importance clinical significance. The model preserved the monotonic changes, while maintaining the quality of reconstruction. In addition, the model provided the capability to generate random samples that satisfy the morphological characteristics at any point of the pathology continuum.

A crucial challenge in image-based modeling of biomedical data is to identify trends and features that separate normality and pathology. In many cases, the morphology of the imaged object exhibits continuous change as it deviates from normality, and thus a generative model can be trained to model this morphological continuum. Moreover, given side information that correlates to certain trend in morphological change, a latent variable model can be regularized such that its latent representation reflects this side information. In this work, we use the Wasserstein Auto-encoder to model this pathology continuum, and apply the Hilbert-Schmitt Independence Criterion (HSIC) to enforce dependency between certain latent features and the provided side information. We experimentally show that the model can provide disentangled and interpretable latent representations and also generate a continuum of morphological changes that corresponds to change in the side information.

\end{abstract}

\section{Introduction}
\label{sec:intro}
{

Machine learning models operating on medical images often aim to identify morphological features that distinguish unhealthy or stressed biological compositions from those that are normal \cite{eulenberg2017reconstructing,esteva2017dermatologist,kraus2017automated}. Many times the morphological features change gradually as the composition deviates from normality (e.g., change of cell shape as concentration of applied drug increases). In such cases, in addition to modeling the data distribution, it would also be useful to explicitly construct features that capture this continuous change and influence the generative process, so that the resulting model has greater interpretability and fidelity. Specifically, provided side information responsible for certain morphological changes, we would like to train a generative model with interpretable latent representation that disentangles this side information, so that the model can generate the corresponding continuum.

Deep generative models have shown great success in modeling medical data \cite{johnson2017generative,mcdermott2018semi}. Typically, a deep generative model learns to transform a prior $P_Z$ into the data distribution $P_X$. While evaluating model likelihood is generally intractable, several good approximations have been proposed, such as optimizing the evidence lower bound (ELBO) \cite{kingma2013auto} or parameterizing an adversarial divergence  \cite{goodfellow2014generative}. An encoder-decoder architecture \cite{bengio2009learning} is useful when inference is required (e.g., given an cell image, encode its morphology into latent features with lower dimensions). In a regular auto-encoder, minimizing the reconstruction cost alone may not result in good latent representations \cite{makhzani2015adversarial}. Therefore, regularizing the encoded latent features is crucial in applying encoder-decoder architecture to generative modeling. The choice of regularizer steers the diversity of interpretations  of the model such as optimizing the ELBO on the data distribution \cite{kingma2013auto}, or minimizing the primal form of Wasserstein distance \cite{tolstikhin2017wasserstein}. 

Side information refers to additional feature that is not directly modeled but may be relevant to the primary task. In few-shot learning or transfer learning, side information is useful in learning robust and generalizable representations \cite{tsai2017learning}. For instance, \cite{tsai2017improving} uses word embedding vectors as side information and applies the HSIC \cite{gretton2005measuring} with learned kernels to enforce its dependency with the learned representation. When dealing with data from source domain and target domain, a classifier can be trained between source and target to encourage features to be domain-invariant\cite{ganin2016domain}. On the other hand, regularization on the individual latent features in encoder-decoder models can also lead to better performance and interpretability. In \cite{chen2018isolating}, the total correlation between latent axes is penalized in order to disentangle the latent features, whereas in \cite{lopez2018information} this regularizer is replace by dHSIC.

In this work, we employ a non-parametric independence measure (HSIC) to integrate side information into the latent representation of a generative model trained on biomedical data. Specifically, given side information that correlates to certain continuous morphological changes of the biological composition, we disentangle the latent features by incorporating this information into one axis, and forcing the remaining axes to be independent to the information. In contrast to classifier-based regularization, this approach does not require training an additional model and thus is more stable and data-efficient. We verify our method on two different biological datasets: lung cancer images acquired by CT scans~\cite{armato2011lung}, and single-cell leukemia images acquired by time-stretch microscopy~\cite{kobayashi2017scirep}. In both experiments, our generative model successfully models continuous morphological changes (side information-dependent) and produces interpretable latent representation that captures the trend.

}

\section{Methods}
\label{sec:method}
{
\paragraph{Wasserstein Auto-encoder}
In this work the generative model is a Wasserstein Auto-encoder \cite{tolstikhin2017wasserstein}. Given the data distribution $P_X$ and generated distribution $P_G$, consider the reparameterized primal form of optimal transport \cite{villani2008optimal, tolstikhin2017wasserstein}:
\begin{align}
W_c(P_X, P_G) = \underset{Q_Z = P_Z}{\mathrm{inf}}\,\mathbb{E}_{P_X}\mathbb{E}_{Q(Z|X)}\Big[c\big(X, G(Z)\big)\Big].
\end{align}
Relaxing $Q_Z = P_Z$, this objective can be written as the minimization of the following loss:
\begin{align}
\calL_{\text{WAE}}(P_{X}, P_{G}) := \mathbbE_{P_X} \mathbbE_{Q(Z|X)}[c(X,G(Z)]
+ \lambda_1 D(Q_Z,P_Z),
\end{align}
in which $D$ is a divergence measure that matches $P_Z$ and $Q_Z$, which we set to be the maximum mean discrepancy (MMD).

\paragraph{Kernel-based Regularizers}

We employ the maximum mean discrepancy (MMD) \cite{gretton2012kernel} to match the prior and aggregated posterior in WAE. The MMD is the RKHS distance between mean embeddings,
and its unbiased empirical estimate can be computed in $O(n^2)$:
\begin{align}
\text{MMD}_u^2(X,Y)
= \frac{1}{m(m-1)} \!\sum_{i \neq j}^m k(\boldx_i,\boldx_j) \!+ \! \frac{1}{n(n-1)}\!\sum_{i \neq j}^n k(\boldy_i,\boldy_j) - \frac{2}{mn} \sum_{i,j}^{m,n} k(\boldx_i,\boldy_j).
\end{align}

With a divergence measure, we can also define an information measure as the discrepancy between the joint distribution and the product of marginal distributions. The Hilbert-Schmitt independence criterion (HSIC) \cite{gretton2005measuring} is defined as the squared MMD between the joint distribution and the product of marginals of two random variables, and its biased empirical estimate is given by:
\begin{align}
\text{HSIC$_{b}$}(X,Y) &= \frac{1}{n^2} \!\sum_{i,j}^n k(\boldx_i,\boldx_j) l(\boldy_i,\boldy_j) \!+\! \frac{1}{n^4}\!\sum_{i,j,q,r}^n k(\boldx_i,\boldx_j)l(\boldy_q,\boldy_r) \notag\\ &- \frac{2}{n^3} \sum_{i,j,q}^n k(\boldx_i,\boldx_j)k(\boldy_i,\boldy_q) = \frac{1}{n^2}\text{tr}\left(\boldK \boldH \boldL \boldH\right),
\end{align}
where $\boldH = \boldI - \frac{1}{n}\boldone \boldone^\top$, $\boldK \in \mathbbR^{n\times n}$ is the Gram matrix of $X$ with $\boldK_{ij} = k(\boldx_i, \boldx_j)$, and $\boldL \in \mathbbR^{n \times n}$ is the Gram matrix of $Y$ with $\boldL_{ij} = l(\boldy_i, \boldy_j)$. Since the time complexity of HSIC is also quadratic, this additional regularizer would increase training time only marginally.

\paragraph{HSIC-regularized WAE}
Given side information $\calS$ we want to incorporate into the latent representation, we can add the following regularizer to the WAE objective to encourage dependence or independence between the aggregated posterior $Q_Z$ and $\calS$:
\begin{align}
\calL_{\text{regularized}} = \calL_{\text{WAE}} + \lambda_2 \text{HSIC}(Q_Z, \calS).
\end{align}
Moreover, we can disentangle the correlation between the side information and certain axis by increasing dependence between $\calS$ and one axis $Z^{dep} = Q_{Z}^{(1)}$ and decreasing the dependence between $\calS$ and the remaining axes $Z^{ind} = Q_{Z}^{(2\sim n)}$: 
\begin{align}
\calL_{\text{regularized}} = \calL_{\text{WAE}}  + \lambda_2 \text{HSIC}(Q^{(2\sim n)}_Z, \calS) - \lambda_3 \text{HSIC}(Q^{(1)}_Z, \calS).
\end{align}
In this case, information of $\calS$ would concentrate at $Z^{dep}$; therefore, by varying this axis we can generate a continuum of morphological changes that correspond to change in $\calS$.

}

\section{Data and Experiments}
\label{sec:experiments}

{
\paragraph{LIDC-IDRI dataset}

The Lung Image Data Consotium (LIDC) \cite{armato2011lung} comprises of thoracic scans from 1018 patient cases with 2670 images. Each sample includes the coordinates contouring the susceptible nodule in the CT scan and radiologists’ assessment of likelihood of malignancy ranged from 1 to 5. Nodules were extracted and the final image was obtained by taking the union of all nodule. The dimension of input image was 48 x 48. Data was augmented by rotation and reflection.

\paragraph{K562 Cell Image Dataset}
A leukemia cell line K562 was divided and incubated with 10-fold serial dilutions of adriamycin ranging from 0.5 to 500 nM for 24 hours, followed by image acquisition with optofluidic time-stretch microscope  \cite{goda2009serial}\cite{lei2016optical}. Approximately 10,000 single-cell images were acquired for each concentration. Each image was normalized to 0-mean, down-sampled to 96 x 96, and labeled with the treated drug concentration. It has been reported that drug-induced morphological changes of K562 cells can be captured by the microscopy images \cite{kobayashi2017scirep}. 

\paragraph{Experimental Details}

We trained the HSIC-regularized WAEs on both datasets, with the side information $\calS$ being the malignancy score in LIDC-IDRI dataset, and ADM concentration in K562 dataset. We set $c(x,y) =  \Vert x-y \Vert^2$ and $P_Z$ as a factorized unit Gaussian $p(\boldx) = \frac{1}{\sqrt{(2\pi)^d}}e^{-\frac{\boldx^\top\boldx}{2}}$. For MMD we used the inverse-multiquadratic (IMQ) kernel $k(\boldx,\boldy)  = \frac{1}{\sqrt{\Vert \boldx-\boldy\Vert^2_2 + 1}}$, and for HSIC we used the Gaussian RBF kernel $k(\boldx,\boldy) =  e^{-\frac{\Vert \boldx - \boldy\Vert_2^2}{2\sigma^2}}$ with bandwidth selected via the median trick. We used Adam \cite{kingma2014adam} for optimization. Implementation details are included in the appendix.
}

\section{Results}
\label{sec:results}
{
\begin{figure}[h!]
\minipage{0.43\textwidth}
  \includegraphics[width=1.1\linewidth]{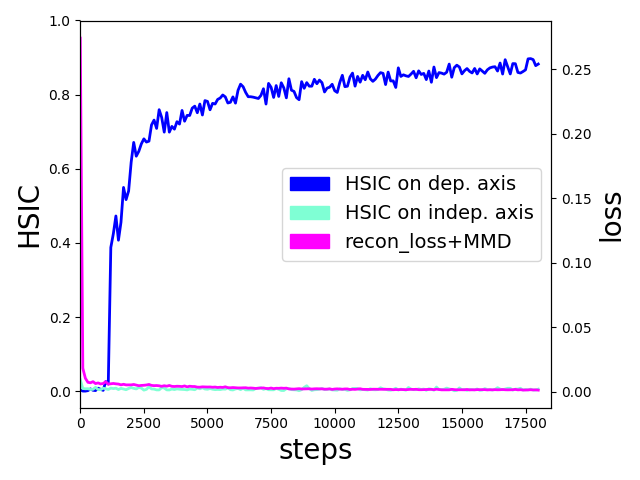}
%   \caption{reconstruction loss}
\endminipage\hfill
\minipage{0.46\textwidth}
  \includegraphics[width=\linewidth]{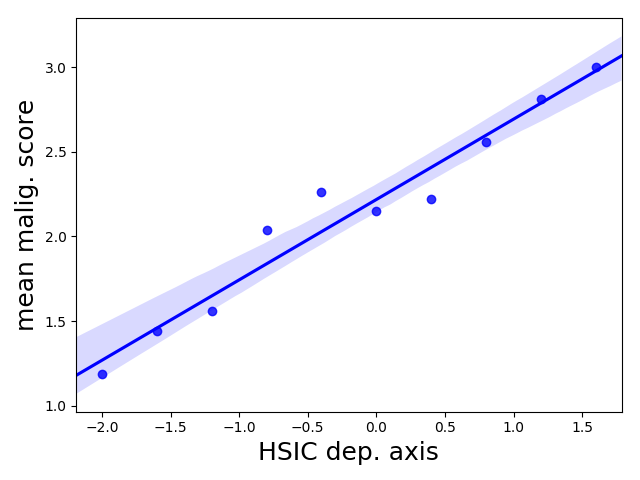}
%   \caption{mmd loss}
\endminipage\hfill
\caption{Results from LIDC-IDRI dataset. Left: training loss and HSIC loss vs. training steps. Right: malignancy score of the nearest neighbors of generated samples vs. $Z^{dep}$; the trend of malignancy correlates with the dependent axis.}
\label{fig:loss}
\end{figure}
\subsection*{HSIC disentangles latent features with respect to side information}
Training loss for the LIDC-IDRI dataset is shown in Figure~\ref{fig:loss}. It can be observed that in addition to minimizing the WAE loss, training also pushes the HSIC($Z^{ind},\calS$) towards 0, indicating that these axes contain little to no information of the label, and at the same time consistently increases HSIC($Z^{ind},\calS$). To further verify that this dependency is captured by the model, we generated images from random $Z$ (see Appendix for generated images) and found their 3 nearest neighbors in the test data; we then regressed the malignancy score $\calS$ of these nearest neighbors against the dependent axis $Z^{dep}$. The regression plot in Figure ~\ref{fig:loss} suggests a strong positive increasing trend of malignancy score, thus indicating that trend in $Z^{dep}$ does indeed match the increasing trend of malignancy score. 

\subsection*{Latent representation and generated samples are consistent with side information}

For the K562 dataset, we encode the test images and visualize the latent space via a scatter plot, in which the x-axis is the dependent axis $Z^{dep}$, and the y-axis is the 1st principle component of the independent axes $Z_{ind}$. Consistent with our expectation, Figure~\ref{fig:hiscplot} shows that concentration of the encoded cell images vary dramatically along $Z^{dep}$, but not in the other axes. This observation is further supported by the kernel-fitted densities for each class: $Z^{dim}$ exhibits reasonable separation between different drug concentrations; in contrast, different classes are almost indistinguishable from $Z^{ind}$. Random samples are also generated, and it can be observed that cells seem to become larger in size as the concentration of adriamycin increases. This finding agrees with the cellular mechanism: adriamycin can arrest cells in G2/M phase just before mitosis, and thus the druge-affected cells tends to be larger in volume \cite{giuseppe1989adriamycin}. Meanwhile, morphological changes other than size change are also present in the manifold, suggesting unelucidated features to be investigated in further studies.

\begin{figure}[h!]
    \centering
    \includegraphics[width=0.98\linewidth]{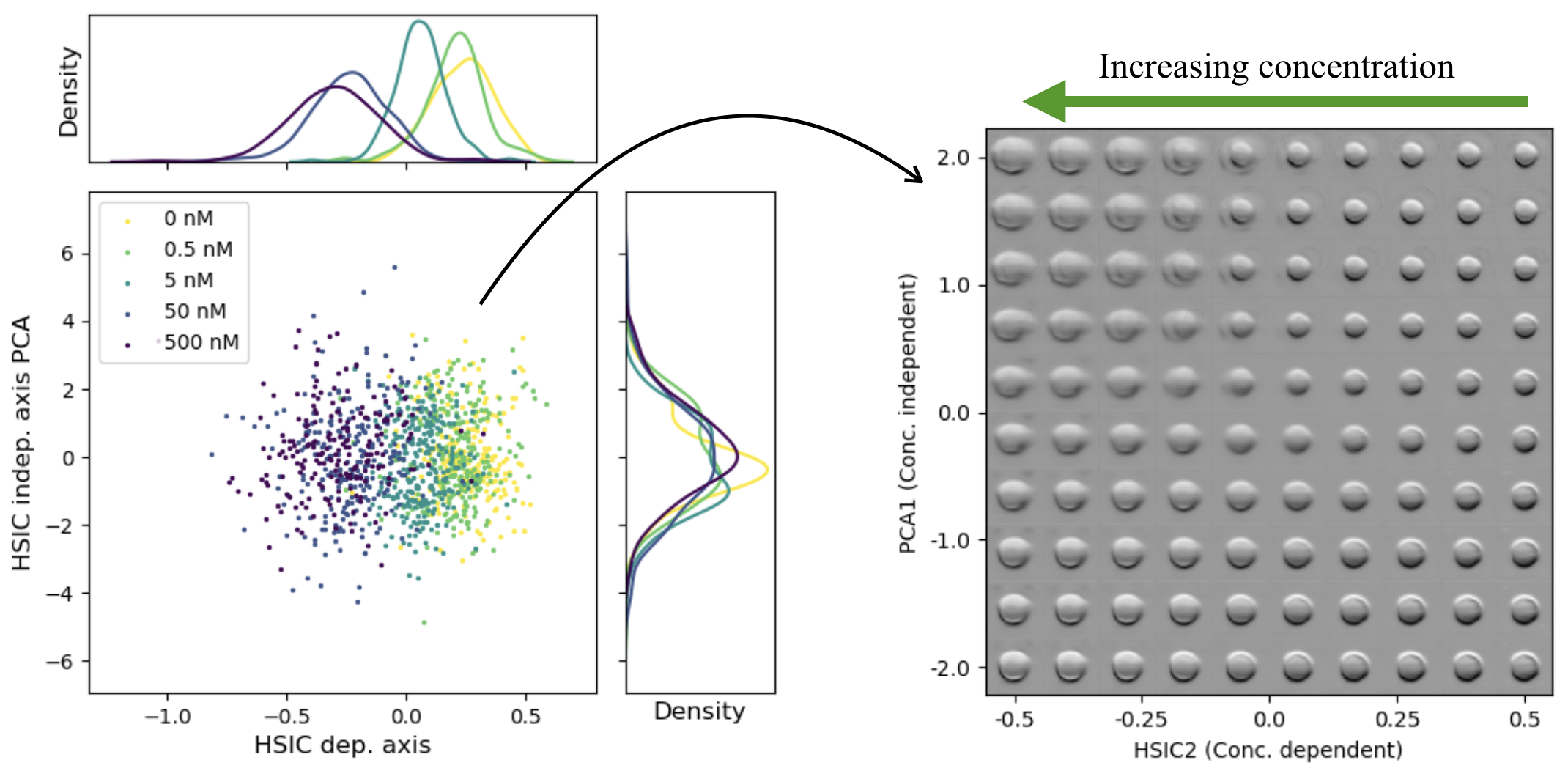}
    \caption{Results from K562 dataset. Left: scatter plot of test images on latent space, with $Z^{dep}$ as x-axis and the 1st principle component (PC) of $Z^{ind}$ as y-axis; class separation is obvious on $Z^{dep}$ but not on other axes. Right: generated images sampled from the dependent axis and the 1st PC of all other axes; generated cells vary in shape along $Z^{dep}$.}
    \label{fig:hiscplot}
\end{figure}

}

\section{Conclusion}

In this work we proposed a regularized generative model that constructs interpretable latent features and models continuous morphological change that corresponds to the provided side information. We applied our model to two distinct biomedical datasets with different clinical significance and validated its effectiveness in incorporating and disentangling the side information in the latent representation as well as generation, which enables modeling of a continuous spectrum of morphological changes.

\newpage
{\footnotesize
\bibliographystyle{unsrt}
\bibliography{main}
}

\newpage

\section{Appendix}
\label{sec:appendix}
{
\subsection{Further details on experiments}

\subsubsection*{Model hyperparameters for LIDC-IDRI dataset}

We used batches of size 512 and trained the models for 18000 steps, approximately 5 epochs. We used $\lambda_1 = 1$, $\lambda_2 = 0.002$, $\lambda_3 = 0.05$ and $\sigma^2_z = 1$. We set $\alpha = 10^{-3}$ for Adam optimizer.

Encoder architecture:
\begin{align*}
\begin{split}
x \in \mathbbR^{48x48} &\xrightarrow{} Conv_{64} \xrightarrow{} BN \xrightarrow{} LReLU \xrightarrow{} MAXPOOL\\
&\xrightarrow{} Conv_{128} \xrightarrow{} BN \xrightarrow{} LReLU \xrightarrow{} MAXPOOL\\
&\xrightarrow{} Conv_{256} \xrightarrow{} BN \xrightarrow{} LReLU\\
&\xrightarrow{} FC_{256} \xrightarrow{} LReLU \xrightarrow{} FC_{10}
\end{split}
\end{align*}

Decoder architecture:
\begin{align*}
\begin{split}
z \in \mathbbR^{10} &\xrightarrow{} FC_{256} \xrightarrow{} LReLU
\xrightarrow{} FC_{3*3*256} \xrightarrow{} LReLU\\
&\xrightarrow{} FSConv_{128} \xrightarrow{} BN \xrightarrow{} LReLU\\
&\xrightarrow{} FSConv_{64} \xrightarrow{} BN \xrightarrow{} LReLU\\
&\xrightarrow{} FSConv_{32} \xrightarrow{} BN \xrightarrow{} LReLU \\ &\xrightarrow{} FSConv_{1} \xrightarrow{} SIGMOID
\end{split}
\end{align*}

Where $Conv_k$ stands for the convolutional layer with k filters, $FSConv_k$ for the fractional strided convolution layer with k filters, $BN$ for the batch normalization, $LReLU$ for the leaky rectified linear units and $FC_k$ for the fully connected layer. All the convolutional layers in the encoder and decoder used vertical and horizontal strides of 2 and SAME padding.

\subsubsection*{Model hyperparameters for K562 Dataset}
For training HSIC-regularized WAE on K562 dataset, we used batches of size 200, and trained the models for 8000 steps, approximately 50 epochs. We used $\lambda_1 = 10$, $\lambda_2 = 0.2$, $\lambda_3 = 0.01$ and $\sigma^2_z = 1$. We set $\alpha = 10^{-3}$ for Adam optimizer.

Encoder architecture:
\begin{align*}
\begin{split}
x \in \mathbbR^{96x96} &\xrightarrow{} Conv_{16} \xrightarrow{} BN \xrightarrow{} LReLU \xrightarrow{} MAXPOOL\\
&\xrightarrow{} Conv_{16} \xrightarrow{} BN \xrightarrow{} ReLU \xrightarrow{} MAXPOOL\\
&\xrightarrow{} Conv_{32} \xrightarrow{} BN
\xrightarrow{} ReLU \xrightarrow{} MAXPOOL\\
&\xrightarrow{} Conv_{32} \xrightarrow{} BN
\xrightarrow{} ReLU \xrightarrow{} MAXPOOL\\
&\xrightarrow{} FC_{32}\\
\end{split}
\end{align*}

Decoder architecture:
\begin{align*}
\begin{split}
z \in \mathbbR^{32} &\xrightarrow{} FC_{6*6*32} \xrightarrow{} ReLU \\
&\xrightarrow{} FSConv_{64} \xrightarrow{} BN \xrightarrow{} ReLU \\
&\xrightarrow{} FSConv_{64} \xrightarrow{} BN \xrightarrow{} ReLU \\
&\xrightarrow{} FSConv_{64} \xrightarrow{} BN \xrightarrow{} ReLU \\
&\xrightarrow{} FSConv_{1} \xrightarrow{} LINEAR
\end{split}
\end{align*}

\subsection{Additional Results}

\begin{figure}[H]
\minipage{0.45\textwidth}
  \captionsetup{labelformat=empty}
  \includegraphics[scale=0.35]{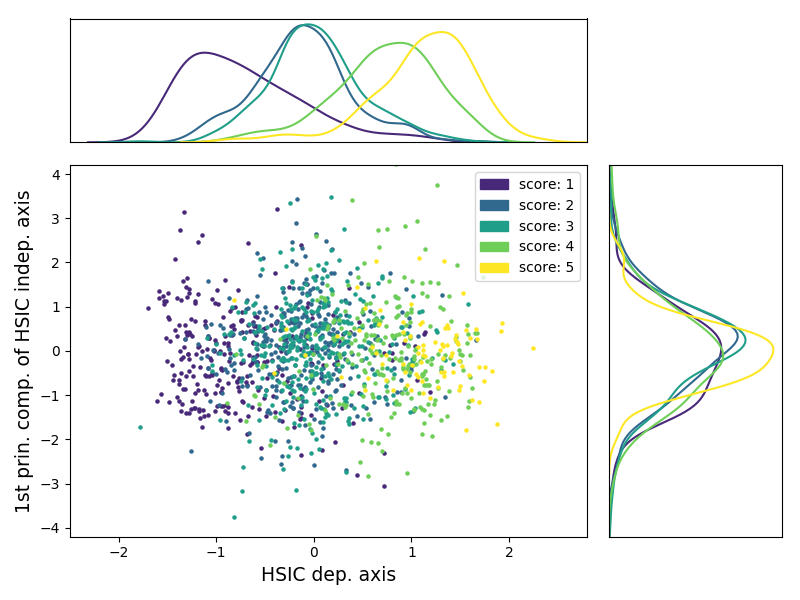}
\endminipage\hfill
\minipage{0.48\textwidth}
  \captionsetup{labelformat=empty}
  \includegraphics[scale=0.22]{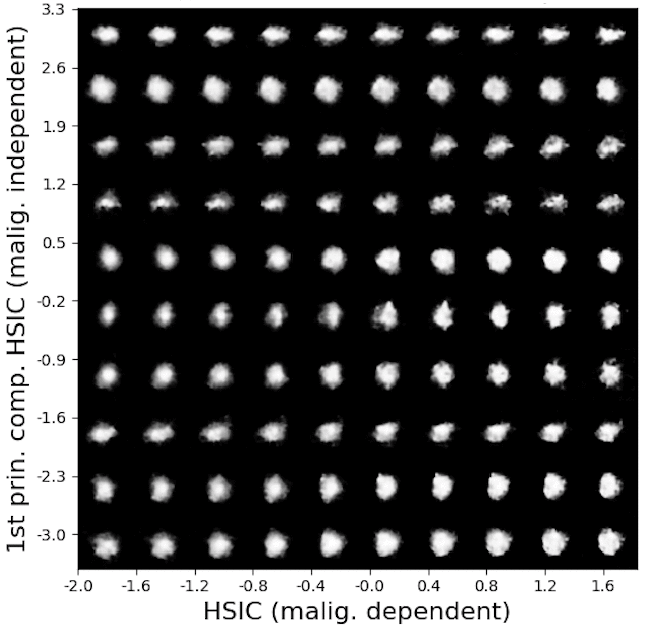}
\endminipage\hfill
\caption{Additional results from LIDC-IDRI dataset. 
Left: scatter plot of test images on 10D latent space; x-axis is the feature dependent to the malignancy score information, and y-axis the 1st principle component (PC) of all independent features. Consistent with our expectation and our observation from the K562 cell image dataset, Figure~\ref{fig:sample} shows that malignancy of lung cancer nodule in CT scans vary dramatically along HSIC dep. axis, while not in the other axes. The kernel density estimation plot for each malig. score group also shows strong separations between the distributions along the HSIC dep. axis, while little effects were contributed by other axis, resulting in indistinguishable distributions. 
Right: The generated samples from HISC-regulated MMD WAE's decoder identifies that the morphological change seems to be the principle feature that influence the malignancy score of lung cancer nodules in CT scan. The nodules' shape becomes pointier, less rounder, and the pixel of the image intensifies as the malignancy score increases.}
\label{fig:sample}
\end{figure}
}

\end{document}